\documentclass[a4paper, 10pt, conference]{ieeeconf}      % Use this line for a4
\IEEEoverridecommandlockouts                              % This command is only
\usepackage[utf8]{inputenc}
\usepackage[T1]{fontenc}
\usepackage{amsmath}

\usepackage[linesnumbered,vlined,ruled,boxed,commentsnumbered]{algorithm2e}
\usepackage{graphicx}
\usepackage{booktabs}
\usepackage{breqn}
\usepackage[acronym]{glossaries}
\usepackage{subfig}
\usepackage[dvipsnames]{xcolor}

\title{\LARGE \bf

An analysis of Reinforcement Learning applied to Coach task in IEEE Very Small Size Soccer
}

\author{
    Carlos H. C. Pena$^{1}$, Mateus G. Machado$^{1}$, Mariana S. Barros$^{1}$, José D. P. Silva$^{1}$, Lucas D. Maciel$^{1}$,\\ Tsang Ing Ren$^{1}$, Edna N. S. Barros$^{1}$, Pedro H. M. Braga$^{1}$, Hansenclever F. Bassani$^{1}$
    \thanks{$^{1}$Centro de Informática - Universidade Federal de Pernambuco, Av. Jornalista Anibal Fernandes, s/n - CDU 50.740-560, Recife, PE, Brazil. Corresponding author: Carlos Pena {\tt\small chcp@cin.ufpe.br}}

}

\begin{document}

\maketitle
\thispagestyle{empty}
\pagestyle{empty}

\begin{abstract}
The IEEE Very Small Size Soccer (VSSS) is a robot soccer competition in which two teams of three small robots play against each other. Traditionally, a deterministic coach agent will choose the most suitable strategy and formation for each adversary's strategy. Therefore, the role of a coach is of great importance to the game. In this sense, this paper proposes an end-to-end approach for the coaching task based on Reinforcement Learning (RL). The proposed system processes the information during the simulated matches to learn an optimal policy that chooses the current formation, depending on the opponent and game conditions. We trained two RL policies against three different teams (balanced, offensive, and heavily offensive) in a simulated environment. Our results were assessed against one of the top teams of the VSSS league, showing promising results after achieving a win/loss ratio of approximately 2.0.
\end{abstract}

\newlength\mylen
\newcommand\myinput[1]{%
  \settowidth\mylen{\KwIn{}}%
  \setlength\hangindent{\mylen}%
  \hspace*{\mylen}#1\\}

\let\oldnl\nl% Store \nl in \oldnl
\newcommand{\nonl}{\renewcommand{\nl}{\let\nl\oldnl}}% Remove line number for one line
  
\newcommand{\fref}[1]{Fig.~\ref{#1}}
\newcommand{\tref}[1]{Table~\ref{#1}}
\newcommand{\crefrangeconjunction}{--}
\newcommand{\sref}[1]{Section~\ref{#1}}
\newcommand{\eref}[1]{Eq.~\ref{#1}}
\newcommand{\aref}[1]{Algorithm~\ref{#1}}

\newacronym{vss}{VSSS}{IEEE Very Small Size Soccer}
\newacronym{pid}{PID}{proportional integral derivative}
\newacronym{rl}{RL}{Reinforcement Learning}
\newacronym{ac}{AC}{Actor-Critic}
\newacronym{isvm}{ISVM}{Improved Support Vector Machine}
\newacronym{svm}{SVM}{Support Vector Machine}
\newacronym{adma-rl}{ADMA-RL}{Adaptive Decision-Making Algorithm
using RL}
\newacronym{pso}{PSO}{Particle Swarm Optimization}
\newacronym{ddpg}{DDPG}{Deep Deterministic Policy Gradient}
\newacronym{dqn}{DQN}{Deep Q-Network}
\newacronym{ddqn}{DDQN}{Double Deep Q-Network}
\newacronym{larc}{LARC}{Latin  American  Robotics  Competition}

\section{Introduction}
\label{sec:intro}

\begin{figure}[thpb]
	\centering
	\includegraphics[width=0.7\linewidth]{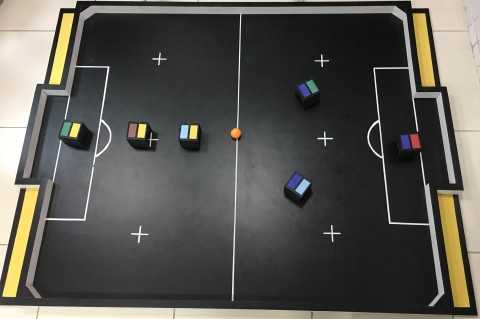}
	\caption{IEEE Very Small Size Soccer.}
	\label{fig:vss}
\end{figure}

The \gls{vss} is a robot soccer competition, in which the teams are composed of three autonomous robots. During the game, the main task for each team  is to navigate on the field, moving the ball to score goals. Once the match starts, the robots cannot receive any human interaction, except during the break time. An image of the \gls{vss} field is shown in Fig.~\ref{fig:vss}. 

\begin{figure}[thpb]
	\centering
	\includegraphics[width=0.4\linewidth]{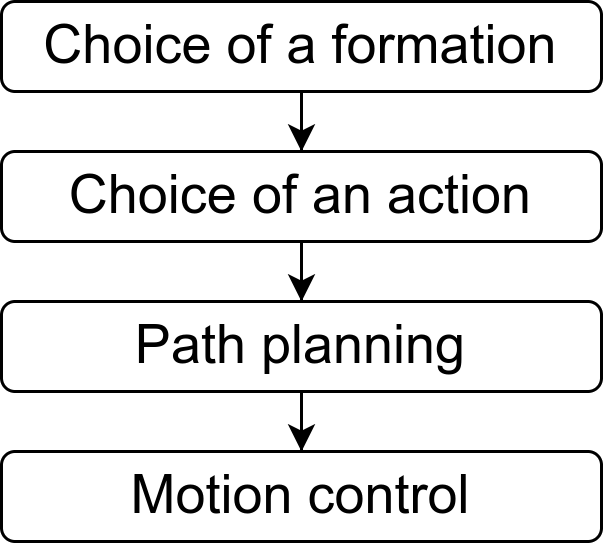}
	\caption{Strategy Module.}
	\label{fig:control}
\end{figure}

Figure~\ref{fig:control} shows a typical architecture of a strategy module for robot soccer. First, given the camera's processed data, such as robots and ball positions and velocities, the strategy module chooses the team formation by defining the role of each robot (e.g., attacker, defender or goalkeeper). In the next step, a target point is assigned for each robot. Path planning algorithms \cite{hart1968formal, zhou2011b,  li2003design} calculate the best path for the robot to navigate, avoiding obstacles. The last step is the motion control, in which the wheel speeds for each of the robot is calculated based on the target point given by the path planning algorithm \cite{kim2004soccer}. 

Recently, with the competition development and the teams' evolution, the games are getting more competitive each year. Therefore, choosing the formation to be used in the different situations of the matches is an essential part of the team strategy. The decision of the role to be played by each robot is essential and may decide the outcome of the match. 

Given the importance of this choice to the team's strategy, this paper approaches this problem and focuses on the first step shown in \fref{fig:control}, which may be called a "coach task". The system gets the information from the game and outputs the optimal formation given the current match situation. In this work, we employed a Reinforcement Learning algorithm in order to train a policy that learns to select a suitable formation in real-time. 

The rest of this article is organized as follows. Section \ref{sec:related} presents the related works for Coach tasks applied to robot soccer. Section \ref{sec:rl} explains and describes the structure of a Reinforcement Learning algorithm. Section \ref{sec:problem} defines the main problem being solved. Section \ref{sec:meth} describes the proposed approach. Section \ref{sec:exp} details the experiments performed. Section \ref{sec:res} presents the results from these experiments and analyses them. Section \ref{sec:conclusion} draws the main conclusions of the article.

\begin{figure}[!ht]
    \centering
     {%
       \includegraphics[width=0.4\textwidth]{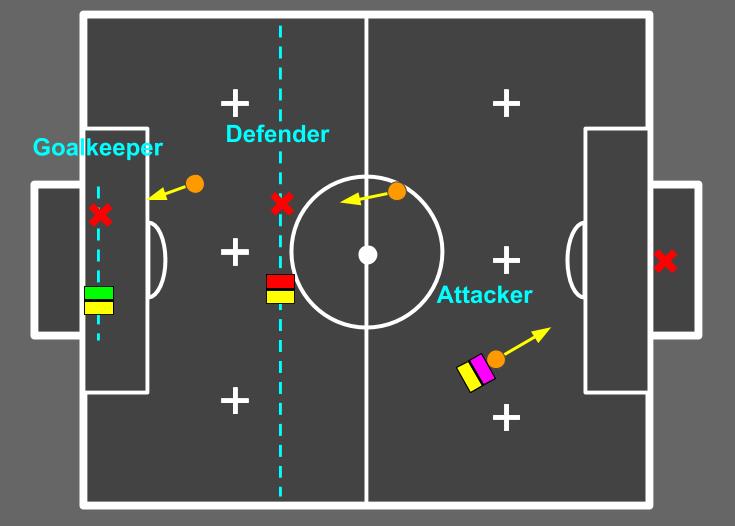}
     }
    
     \caption{Example of the player's roles (goalkeeper, defender and attacker), where the red "X" is the target position for the given robot.}
     \label{fig:lines}
   \end{figure}

\section{Related Work}
\label{sec:related}

In \cite{shi2018fuzzy}, it is proposed a system based on Fuzzy Bayesian \gls{rl} applied to a robotic soccer team. The system is composed of two parts: situation evaluation and decision-making. In the first part, the method extracts information from the soccer games (e.g., remaining time, score difference, and ball distance). These features are used as inputs for a situation evaluation algorithm that outputs the competition situation, among a set of five gradual labels that describe if the match is advantageous for one team or another. In the second part of the method, the previous output is combined with other information, such as ball position and the difference of distance from both teams to the ball, to calculate the probability of the team adopting an offensive, defensive or balanced attitude. Then, a decision-making method with Bayesian \gls{rl} network calculates the best decision. The proposed method is compared with a Fuzzy neural network and a Bayesian-SOM neural network and tested against three different opponents: an aggressive team, a defensive team, and a standard balanced team.%\footnotemark[\ref{note1}]

Moreover, \cite{shi2018adaptive} also proposes an adaptive decision-making method, in which an \gls{isvm} is used to collect and classify the environment and situational information, whereas the \gls{adma-rl} chooses the proper strategy adaptively. First, some evaluation factors are used as features to represent the situations in the match. Then, it performs the Situation Classification, a process that combines an \gls{svm} algorithm with a \gls{pso} to find the optimal solution, using a Decision Tree to execute the multi-classification. In sequence, the adaptive Decision-Making method uses a Q-learning algorithm to learn the best actions to be taken at each state. At this point, the state represents the match scenario, and the actions are the possible strategies for the team to choose. The proposed method was compared to a decision-making technique using a fuzzy neural network and another one that uses a Bayesian-SOM neural network.%\footnote{\label{note1}Some information in \cite{shi2018fuzzy} and \cite{shi2018adaptive}, such as the supervised dataset and shooting ability, could not be directly adapted to the \gls{vss} environment.} %Although we could provide a dataset based on the simulator used by us, it could reproduce unfaithful results.}

Recently, \cite{bassani2020framework} proposed a framework to study \gls{rl} in simulated and real environments. As we will show in \sref{sec:sim}, our approach is based on theirs. They modified the FIRASim \cite{fira} and VSS-SDK \cite{vss_sdk} simulators to communicate with an OpenAI Gym \cite{gym} environment, and control each wheel speed from each robot. The chosen simulator communicates with a Gym environment that interacts with the agents. Their research shows great results for both single and multi-agent environments. Our approach is similar, but our task is about deciding which role to choose, given a set of states (the first step in Fig.
~\ref{fig:control}), rather than how to control wheels' speeds given a state (the other steps in Fig.
~\ref{fig:control}).

\section{Reinforcement Learning}
\label{sec:rl}

\begin{figure}[!ht]
    \centering
    \includegraphics[width=0.9\linewidth,scale=1]{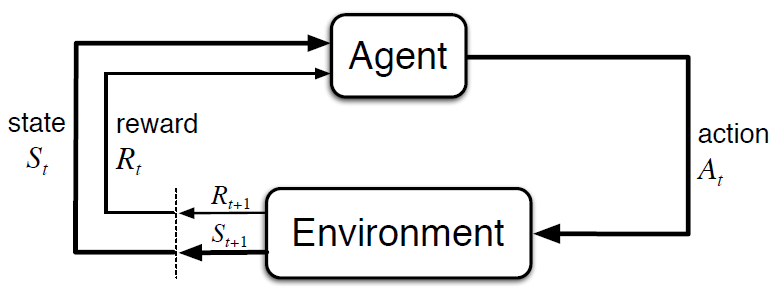}
    \caption{Interaction between agent and environment in a RL method \cite{Sutton}.}
    \label{fig:rl}
\end{figure}

Reinforcement Learning (RL) can be described as the sub-area of Machine Learning in which the agent learns by trial-and-error, while it interacts with the environment \cite{Sutton}. At each state, the agent will perform an action in the environment and receive from it a reward signal that informs how good or bad are the actions taken so far, as illustrated in \fref{fig:rl}. In an RL method, the policy defines the agent behavior at each state depending on the environment, so the agent's main objective is to find a policy that maximizes the expected cumulative total reward, usually represented by a value function \cite{Sutton}.

A very successful algorithm in RL is the Q-learning \cite{qlearning}. In this method, the value function is updated based on the Bellman equation, and the value of the optimal policy is learned independently of the agent's actions. The main idea is to learn the quality, or how useful a given action is, in a particular state, in yielding future rewards. The cumulative discount reward is the expected value of choosing a specific  action in the current state while following the current optimal policy. An action is chosen in the algorithm loop, and then the respective Q-value is updated using the Bellman equation \eqref{bellmaneq}. This process is repeated until the convergence of the value function. 

\begin{equation}
    V(s) = \max_{a}(R(s,a) + \gamma*V(s')) 
    \label{bellmaneq}
\end{equation}

A Q-learning method that usually achieves high performance is \gls{dqn}. It is essentially a Q-learning technique that uses a deep neural network \cite{dqn} to estimate the action value function (Q-Value). Therefore, the inputs of the \gls{dqn} are the states, and the outputs are the estimated optimal values of each action in this state. \gls{dqn} achieves excellent results, especially in applications related to games \cite{dqn} \cite{starcraft}. One crucial characteristic of this method is the experience replay that stores the agent's experience at each time step. This buffer is used in order to perform the network's weights updates using Stochastic Gradient Descent (SGD).

As shown in \cite{ddqn}, the \gls{dqn} produces overly optimistic Q-Values comparing with the real ones. With that behavior, the Neural Network can converge to a local optimum. The \gls{ddqn} addresses the overestimation of Q-Values, by adding a target network that will produce Q-Values based on \eqref{eq:ddqn}, where $\theta' $ is the target network, and the online network will receive a loss signal based on \eqref{eq:ddqnloss}. After $\tau$ steps, the target network should be updated with the online network. The main idea is to maximize the loss if $Q(S, \theta) >> Q(S, \theta')$.

\begin{equation}
    Y_t = R_{t+1} + \gamma\text{ argmax}(Q(S_{t+1}, \theta'))
    \label{eq:ddqn}
\end{equation}

\begin{equation}
    \mathcal{L}_{\theta} = ||Q(S, \theta) - Y_t||^2
    \label{eq:ddqnloss}
\end{equation}

\begin{figure}[!ht]
     \centering
     \includegraphics[width=0.6\linewidth,scale=1]{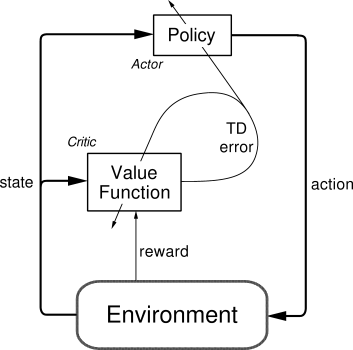}
     \caption{Actor-Critic training basic architecture \cite{Sutton}.}
     \label{fig:acmethods}
 \end{figure}

 Another successful family of approaches for \gls{rl} are the \gls{ac} methods. They learn approximations to both the policy and the value functions. In this case, ``actor'' refers to the learned policy parameters, $\phi$, and "critic" refers to the learned value-function parameters, $\theta$. As shown in \fref{fig:acmethods}, the actor uses the critic evaluation as a loss function. The critic is updated as a value function.
 
The \gls{ddpg} is an \gls{ac} method that uses Neural Networks to approximate the Actor and Critic function \cite{ddpg}. The \gls{ddpg} algorithm already incorporates target networks for both Actor and Critic using a soft update technique that will update the target Networks every step, but at a slower rate of $\tau$ discounted on the online Networks weights. The \gls{ddpg} training is like an usual \gls{ac} method, but using Neural Networks. The Critic updates its weights according to \eqref{eq:criticloss1} and \eqref{eq:criticloss2}; and the Actor according to \eqref{eq:actorloss} which $\mu_\phi$ and $\mu_{\phi'}$ are the policy and target policy functions respectively.

\begin{equation}
    Y_t = R_{t+1} + \gamma\text{ argmax}(Q(S_{t+1}, \mu_{\phi'}(S_{t+1}), \theta'))
    \label{eq:criticloss1}
\end{equation}

\begin{equation}
    \mathcal{L}_{\theta} =  ||Q(S, \mu_\phi(S), \theta) - Y_t||^2
    \label{eq:criticloss2}
\end{equation}

\begin{equation}
    \mathcal{L}_{\phi} = -Q(S, \mu_\phi(S), \theta) 
    \label{eq:actorloss}
\end{equation}

\section{Problem Definition}
\label{sec:problem}

An IEEE \gls{vss} category competition game is divided into two halves of 5 minutes. During the game, neither of the teams can manually change the strategy or team formation. An exception is on the break in the middle of the game or on the requested breaks during the match. At these moments, the teams usually change their players' roles or the team's formation based on their understanding of the game. 

Regardless of the team strategy, it should avoid committing faults. In the IEEE \gls{vss}, the principal fault is the penalty, when more than one defender robot is inside the goal area along with the ball.
% \begin{itemize}
%     \item Attack Fault: when more than one attack robot is inside the goal area and the second robot touches the ball or inhibits the goalkeeper movement. 
%     \item Penalty: when more than one defender robot is inside the goal area along with the ball.
% \end{itemize}

Unlike human soccer, the roles can change dynamically during the game, so, for example, the goalkeeper could change to an attacker at any time and vice versa. Besides, it is also possible to vary the behavior while playing the same role. For example, a goalkeeper can be either more defensive and stay inside the goal area, or be more offensive and leave the area. These changes made in the team formation and strategy play an important part and can completely change the match's course. They are related not only to the observation and interpretation of the match by the teams but also to each one's reaction ability. Furthermore, since these changes cause different outcomes for the match, there is no optimal formation for the complete game, neither for half of it. It will depend not only on each group's initial formation but also on the changes that happened during the game.

In our current situation, we already have the players' roles working properly. However, we need to know the best moment to choose each combination of them. In the system described in this paper, we use three basic roles: attacker, defender, and goalkeeper. Section \ref{subsec:roles} describes these roles that compose the team formation, and can be seen on \fref{fig:lines}.

\subsection{Roles}
\label{subsec:roles}

\begin{itemize}
  \item Attacker: The main objective of the attacker can be divided into two main parts: 1) reach the ball, and 2) score goals.
  
  \item Defender: The defender's main objective is to stay between the ball and his team's goal line, repelling the ball from his team's area. Given a target point, the defender uses a \gls{pid} controller to stay on a line that follows the target by its position.
  \item Goalkeeper: The goalkeeper follows the same rule as the defender, except that its line position is nearer to the defensive goal. Unlike in human soccer, our goalkeeper does not hold the ball, only push it like the other players.
\end{itemize}

Therefore, the problem addressed in this paper is to dynamically choose an appropriate role, among those defined above, for each one of the robots, during the game.
\section{Proposed Approach}
\label{sec:meth}

In this section we detail our proposed environment, describing it's architecture (Section \ref{sec:sim}); the state and action spaces (Section \ref{sec:feat}); and the chosen reward shaping (Section \ref{sec:rew}).

\subsection{Architecture}
\label{sec:sim}
As stated in \sref{sec:related}, we took as basis the VSSS-RL framework \cite{bassani2020framework} to create our own architecture. We decided to simulate the environment with FIRASim due to its better physics simulation in relation to the real environment.

FIRASim is an open-source project based on GRSim\cite{grsim} to simulate the \gls{vss} environment using Open Dynamics Engine\cite{ode} when it comes to physics. In VSSS-RL, the FIRASim communicates with an OpenAi Gym-based environment that processes the simulation information to structure the state and then passes it to the agent, which returns an action.

\begin{figure}[!ht]
    \centering
    \includegraphics[width=0.8\linewidth,scale=1]{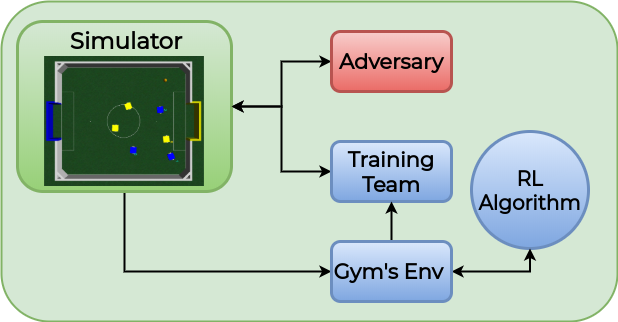}
    \caption{Coach-RL's architecture: High-level actions as strategies to be chosen.}
    \label{fig:architecture}
\end{figure}

We propose the Coach-RL, an architecture that uses two more agent modules to communicate with both the simulator and the environment\footnote{Source code available at https://github.com/RC-Dynamics/Coach-RL}. In order to perform the training, we use a  deterministic agent as an adversary team, and our environment is responsible for starting and stopping the process. The team, trained by a \gls{rl} method, communicates with both environment and simulator, receiving the roles to be chosen at time $\tau$ and sending each robot's motors linear speeds to FIRASim. In \fref{fig:architecture} the complete architecture is illustrated.

\subsection{State and Action spaces}
\label{sec:feat}
Our environment's purpose is to provide strategies for the team. Coach-RL produces as action a combination of the three main roles described in \sref{subsec:roles} for the robots in the team. As there are three robots in a team, it results in twenty-seven discrete available actions for the coach agent to choose. Those actions can be combined in four basic strategies to be chosen: Very offensive, with three attackers; Offensive, with two attackers and one goalkeeper or one defender; Defensive, with two defenders or two goalkeepers and one attacker; Very defensive, with three defenders or goalkeepers
% \begin{itemize}
%     \item Very offensive, with three attackers.
%     \item Offensive, with two attackers and one goalkeeper or one defender.
%     \item Defensive, with two defenders or two goalkeepers and one attacker.
%     \item Very defensive, with three defenders or goalkeepers.
% \end{itemize}

To describe our state-space, we chose the main continuous features of the game: Ball's position, Adversaries' positions; Teammates' positions; and Teammates' angles relative to the field's center ($x=0, y=0$). The coaching task in a football environment is not as dynamic as the player task, which must update commands at every step. A coach should take action based on a stack of frames seen in the last $t-t_1$ timesteps. Based on that premise, we stack each state $N$ times and then act upon our environment.
That allows the \gls{rl} method to learn the dynamics model of change between states, as shown in \cite{predict_fut} \cite{humanlevelcontroldrl}. Another solution would be using a long short-term memory that can, in principle, learn the dynamics model of the environment as in \cite{asyncmethods}. We choose to keep the model engineering simple because this paper's purpose is to put forward a new framework and not a new learning method. 

\subsection{Reward shaping}
\label{sec:rew}
We want our agent to be positively rewarded if the team scores a goal and negatively if it suffers a goal. However, as shown in \cite{bassani2020framework}, goal rewards are not enough for the agent to learn, due to its sparsity, and the reward shaping strategy proposed by the authors was shown to be adequate. 

Therefore, we adopted the gradient of ball's potential presented in \cite{bassani2020framework} in order to reduce the sparsity of the rewards. The reward is described in equations \eqref{eq:potrew} and \eqref{eq:ballpot} where $bp_t$ is the ball potential at step $t$, $d$ is the Euclidean distance between two points, $g_o$ and $g_a$ are each goal center position and $b$ is the ball position. We did not include the rewards proposed in \cite{bassani2020framework} that encourage agents to move towards the ball, since the agent's behaviour already provides this skill.

\begin{equation}
        R_p = \frac{bp_t - bp_{t-dt}}{dt}
\label{eq:potrew}
\end{equation}
\begin{equation}
        bp = \frac{\frac{d(g_o, b) - d(g_a, b)}{d(g_o, g_a)} - 1}{2}
\label{eq:ballpot}
\end{equation}

As we enable extreme strategies, like three goalkeepers or three attackers, we added a negative reward for penalties committed by our players to reduce this kind of infractions. This action may add a bias for a safer game, but, as we will show in the results, as more infractions are made, more goals our agent suffers. The complete reward function is shown below where those values were empirically defined.

\begin{gather*}
Reward =  w_p*R_p +  
\begin{cases}
  +100, & $if it scores a goal$\\
  -35, & $if it makes a penalty$\\
  -100, & $if it concedes a goal$\\
\end{cases}
\end{gather*}
\section{Experiments}
\label{sec:exp}
% - Descrever como a abordagem foi validada
% - Apresentar experimentos feitos
In this section, we provide a description of the experiments developed with Coach-RL using \gls{ddqn} and \gls{ddpg} methods to model our agents, and how we chose to change the adversary strategies to obtain an experienced coach. 
%We could not perform the experiments in a physical environment due to the pandemic situation.

Our approach has a discrete action space that perfectly matches the \gls{ddqn} method. However, the \gls{ddpg} method's output is continuous, so we had to discretize it. For each player, the \gls{ddpg}'s output was divided in three ranges of values: Attacker, if $output < -0.34$; Defender, if $-0.34 < output < 0.34$; Goalkeeper, if $output > 0.34$. Also, we added a time-correlated noise to our \gls{ddpg}'s output, which allows a better exploration of the algorithm.

% \begin{gather*}
% Action =  
% \begin{cases}
%   Attacker, & $if output < -0.34$\\
%   Defender, & $if -0.34 < output < 0.34$\\
%   Goalie, & $if output > 0.34$\\
% \end{cases}
% \label{eq:ddpgout}
% \end{gather*}

We chose the balanced, offensive, and very offensive strategies to play against our training agent. Since the defensive strategies have no attacker in formation, the experiment would not provide fair results, so we chose to keep them out. Every episode, a different strategy is chosen between these three. This training strategy allows a better understanding of what the agent does in each situation and creates a fine-tuned final strategy for teams that can use the three strategies, which is the case of most of \gls{vss}'s teams.

\section{Results}
\label{sec:res}
In this section, we show the results of the performed experiments. Besides, we describe some of the behaviors learned by both algorithms and the comparison with \gls{larc} team.

\subsection{Training Results}
In our training experiments, the agents are evaluated by the accumulated total reward in the window of the next fifteen steps and how many penalties our team commits. We also consider the goal score an important metric, defined by the difference between ours and adversary's goals scored in an episode. 

\begin{figure}[!ht]
    \centering
    \includegraphics[width=1.\linewidth,scale=1]{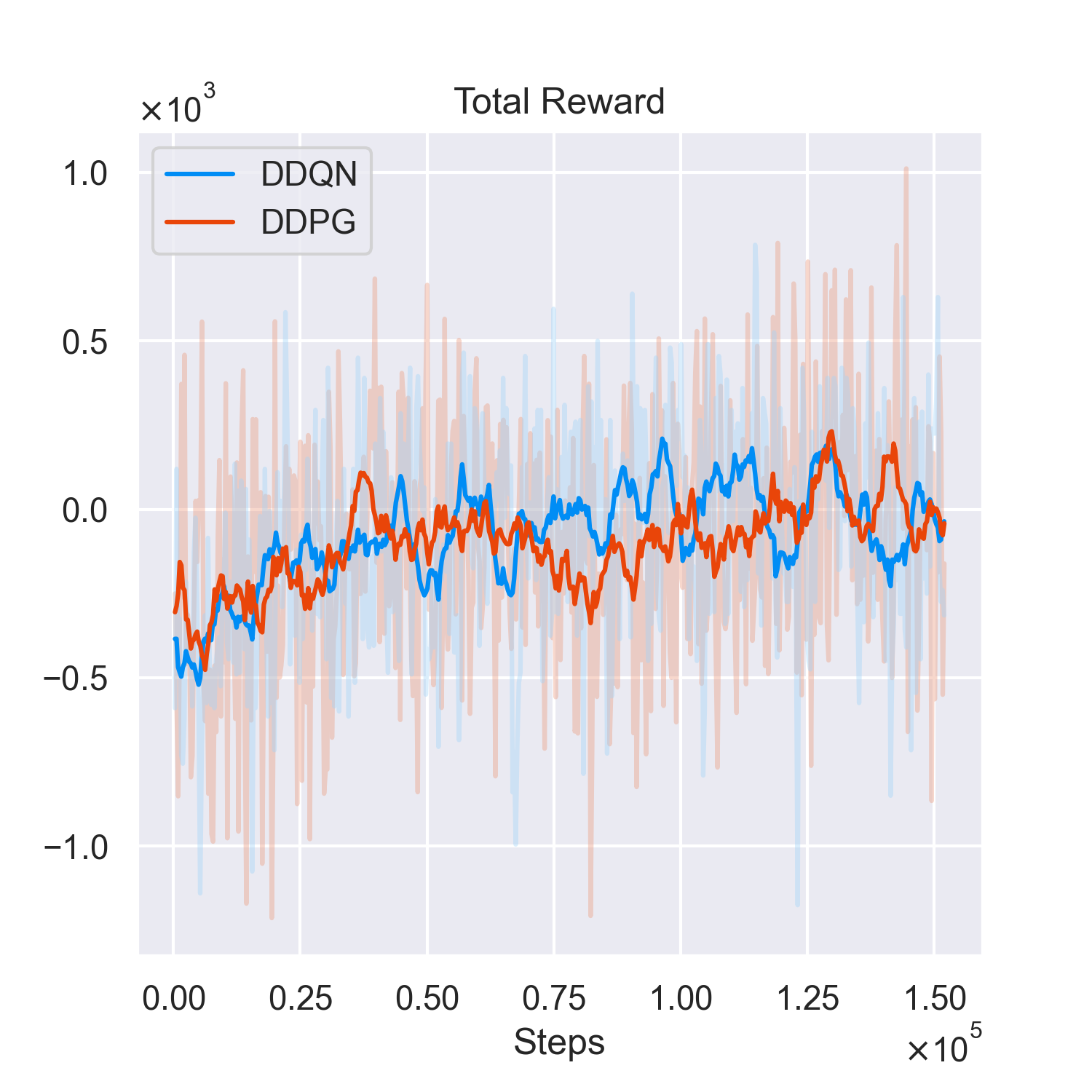}
    \caption{Accumulated rewards along steps for \gls{ddqn} and \gls{ddpg}.}
    \label{fig:goalresults}
\end{figure}

\begin{figure}[!ht]
    \centering
    \includegraphics[width=0.8\linewidth,scale=1]{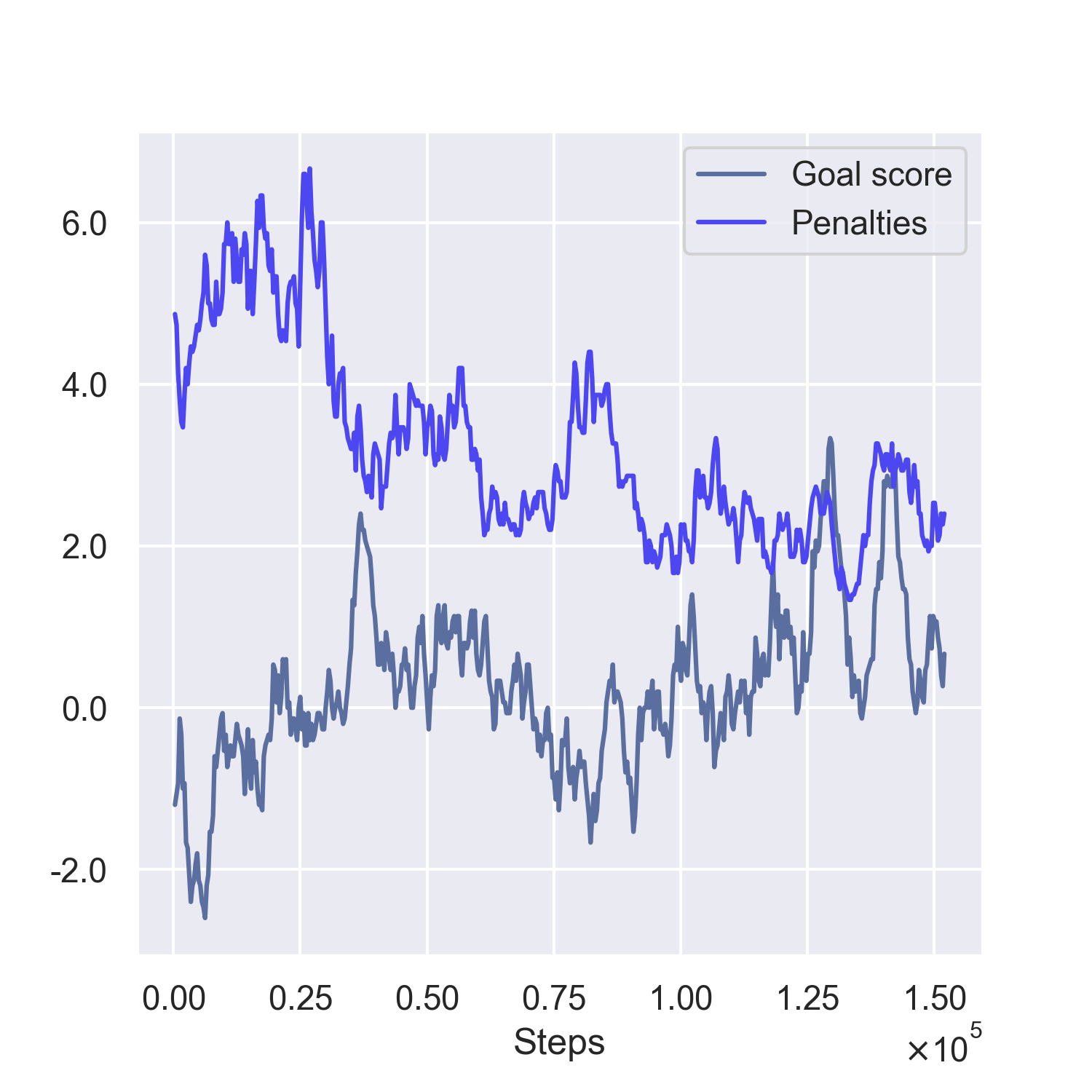}
    \caption{Penalties-Goals score relation for the \gls{ddpg} method.}
    \label{fig:penaltigoal}
\end{figure}

In \fref{fig:goalresults}, it is shown the adaptation of each model during training. Both models presented rising learning curves, indicating that they learn a better strategy for each situation over time. \fref{fig:penaltigoal} shows, as expected, an inverse relation between the goal score and penalties. The more times penalties are committed by the training team, the more goals it suffers. For this reason, we added a negative reward for each time the team commits a penalty.

\begin{figure}[!ht]
    \centering
    \includegraphics[width=1\linewidth,scale=1]{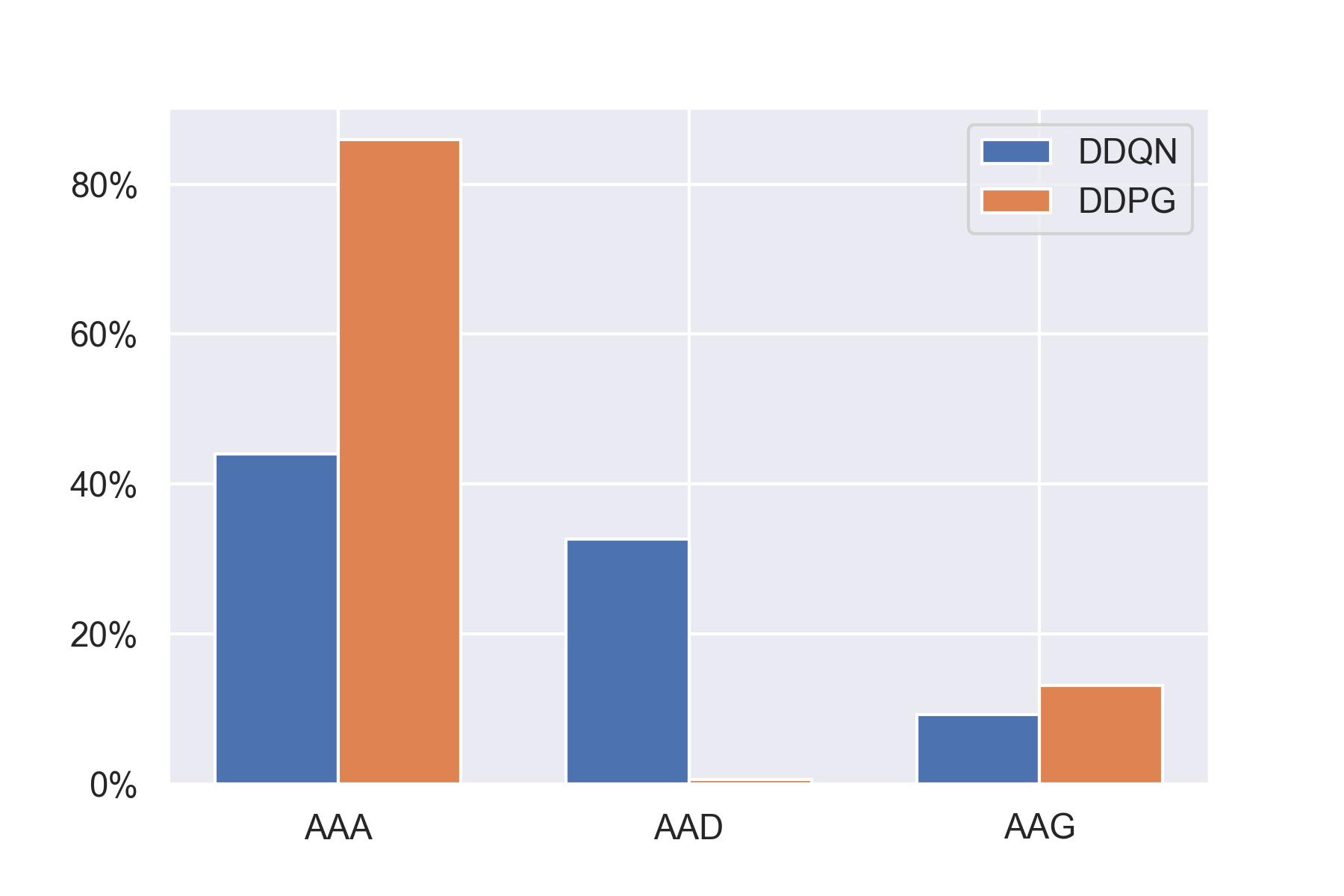}
    \caption{Action distribution percentage of both methods against the different strategies. The X axis means the formation taken, which G is for Goalkeeper, D is for Defender and A is for Attacker. The figure shows only the top 3 chosen strategies.}
    \label{fig:actdist}
\end{figure}

In order to perform the test experiments, the Neural Network weights were frozen. Table ~\ref{tab:vs-det} shows the mean and standard deviation scores for 30 matches between the trained polices and the three opponent strategies used in training. It shows that, although the \gls{ddpg} method lost from the aggressive opponent, it achieved an excellent result against the balanced opponent team. The \gls{ddqn} policy, however, obtained a positive result only against the balanced team.  Analyzing \tref{tab:vs-det} with \fref{fig:actdist}, the trending in our setup is that a balanced strategy is not as useful as an offensive one, what is somehow surprising. The smooth aggressiveness changes in the \gls{ddpg}'s output gives a certain interpretability to the whole output, which helps the system to complete the task better than \gls{ddqn}.

\begin{table}[!ht]
{
\centering
\caption{Mean and std scores for 30 matches, between the proposed models (\gls{ddqn} and \gls{ddpg}) and three static formations (Heavily Aggressive, Aggressive, Balanced)}
\resizebox{0.48\textwidth}{!}{%
\begin{tabular}{ccccc}
\toprule
Policy & Team Score (std)  & & Opponent Score (std) & Opponent \\
\midrule
Coach-RL-DDQN & $2.55 (2.07)$ & $\times$ & $4.17 (2.16)$ & Heavily Aggressive  \\
Coach-RL-DDQN & $2.03 (1.29)$ & $\times$ & $2.75 (1.77)$ & Aggressive  \\
Coach-RL-DDQN & $4.27 (1.52)$ & $\times$ & $1.82 (1.44)$ & Balanced  \\
Coach-RL-DDPG & $3.41 (1.84)$ & $\times$ & $3.41 (1.88)$ & Heavily Aggressive  \\
Coach-RL-DDPG & $2.62 (1.34)$ & $\times$ & $3.06 (1.52)$ & Aggressive  \\
Coach-RL-DDPG & $6.10 (1.86)$ & $\times$ & $1.62 (1.12)$ & Balanced  \\
\bottomrule
\end{tabular}%
}
\label{tab:vs-det}
}

\end{table}

\subsection{Comparison with LARC team}

To evaluate the learned policy, one of the teams that participated in the \gls{larc} 2019 shared with us their binary code. For each policy, we ran 30 episodes, and the score mean and standard deviation can be seen on Table \ref{tab:larc2019}. The invited team has the same strategy module as ours, except for using a deterministic coach algorithm, which selects the formation based on the ball position. Although the \gls{ddqn} policy presents a disadvantaged result, the \gls{ddpg} policy was able to beat the invited team by almost one goal of difference in the mean score. In the final result, \gls{ddpg} won 14 matches and lost 7.

\begin{table}[!ht]
{
\centering
\caption{Mean and std scores of 30 matches, between the proposed models (\gls{ddqn} and \gls{ddpg}) and an team that participate in the \gls{larc}.}
\resizebox{0.48\textwidth}{!}{%
\begin{tabular}{ccccc}
\toprule
Policy & Team Score (std)  & & Opponent Score (std) & Opponent \\
\midrule
Coach-RL-DDQN & $2.51 (1.69)$ & $\times$ & $2.89 (1.39)$ & Team $1$  (5th Place) \\
Coach-RL-DDPG & $3.38 (1.40)$ & $\times$ & $2.48 (1.27)$ &  Team $1$ (5th Place) \\

\bottomrule
\end{tabular}%
}
\label{tab:larc2019}
}

\end{table}
\section{Conclusion}
\label{sec:conclusion}
%- Resumir resultados
%- Principais contribuições
%- Trabalhos futuros

In this work, we propose an end-to-end approach for the Coach task in IEEE \gls{vss}, in which the trained \gls{rl} policy chooses the most suitable formation, depending on the opponent and game conditions. Our experiments show that the \gls{ddpg} policy outperforms the team that obtained fifth place in \gls{larc} 2019. Our evaluation over each trained formation shows a restriction in the low-level actions taken by each player, which leaves room for improvements.

In most experiments, the policy tends to select three attackers more frequently, and often changes to two attackers and one goalkeeper or defender. Therefore, we hypothesize that in the chosen strategy module, the attacker surpasses the other roles, due to the complexity of the attacker algorithm, which includes sub-tasks such as ball control, wall avoidance, and counter-attack. 

For future works, we aim to add more diversity to the set of pre-defined formations, vary the controller settings and the maximum robot speed, and focus on expanding the types of attacker behaviors. Moreover, we intend to evaluate multi-agent methods related to skills learning and a hierarchical approach, in which the coach controls the team's offensiveness and communicates with the three agents trained in a setup  similar to the one described in \cite{bassani2020framework}.

\section*{ACKNOWLEDGMENT}
The authors would like to thank RobôCIn - UFPE Team for the collaboration and resources provided, and Fundação de Amparo a Ciência e Tecnologia do Estado de Pernambuco (FACEPE) for financial support.

\addtolength{\textheight}{-12cm}  

%\section*{APPENDIX}
%Appendixes.

% uncomment for final subimission
% \section*{ACKNOWLEDGMENT}
% ACKNOWLEDGMENT

\bibliographystyle{IEEEtran}
\bibliography{references.bib}
%\end{thebibliography} ? 
\end{document}